\documentclass{article}
\usepackage{spconf,amsmath,graphicx,hyperref}

\usepackage{booktabs} 
\usepackage{multirow} 
\usepackage{array}  
\usepackage{adjustbox}

\usepackage{graphicx} 
\usepackage{subcaption} 
\usepackage{caption} 

\title{Cross-domain EEG-based Emotion Recognition\\ with Contrastive Learning}
\name{Rui Yan$^1$, Yibo Li$^1$, Han Ding$^2$, Fei Wang$^{1*}$\thanks{$^*$ This paper was accepted by IEEE ICASSP 2026. This work was supported by the National Natural Science Foundation of China under Grant Nos. 62572383 and 62372365. The corresponding author is Fei Wang. }}
\address{
$1$ School of Software Engineering, Xi'an Jiaotong University, China\\
$2$ School of Computer Science and Technology, Xi'an Jiaotong University, China \\
\textit{\{yany,ybdeparture\}@stu.xjtu.edu.cn, \{dinghan,feynmanw\}@xjtu.edu.cn }
}

\begin{document}
%
\maketitle
\begin{abstract}
Electroencephalogram (EEG)-based emotion recognition is vital for affective computing but faces challenges in feature utilization and cross-domain generalization. This work introduces EmotionCLIP, which reformulates recognition as an EEG-text matching task within the CLIP framework. A tailored backbone, SST-LegoViT, captures spatial, spectral, and temporal features using multi-scale convolution and Transformer modules. Experiments on SEED and SEED-IV datasets show superior cross-subject accuracies of 88.69\% and 73.50\%, and cross-time accuracies of 88.46\% and 77.54\%, outperforming existing models. Results demonstrate the effectiveness of multimodal contrastive learning for robust EEG emotion recognition. The code is available at \href{https://github.com/Departure2021/EmotionCLIP}{https://github.com/Departure2021/EmotionCLIP}.
\end{abstract}
\section{Introduction}
\label{sec:intro}

Emotion recognition is vital in mental health monitoring, early diagnosis of brain disorders, and enhancing human-computer interaction. Among different modalities, EEG-based methods have become mainstream due to their non-invasiveness, high temporal resolution, and resistance to deliberate manipulation~\cite{pillalamarri2025review}. Compared with facial expressions, speech, or gestures, EEG signals directly reflect brain activity associated with emotions, enabling reliable applications in healthcare, intelligent driving, and interactive systems.

Early methods relied on handcrafted features with support vector machine~\cite{wang2011eeg} or k-nearest neighbor classification~\cite{bahari2013eeg}, which required expert knowledge and were sensitive to noise. Later, deep learning enables end-to-end feature learning and has significantly advanced EEG-based emotion recognition. Convolutional neural networks~\cite{deng2021sfe} captures local dependencies, and graph convolutional neural networks~\cite{song2018dgcnn} and spatial-temporal recurrent neural networks~\cite{zhang2018spatial} model EEG’s topology and dynamics. Recently, Transformer-based models \cite{zhai2025one}, such as compact convolutional Transformers~\cite{song2022eegconformer} and multidimensional attention models~\cite{xu2023amdet}, further improved performance by capturing spatio-temporal-frequency features.



Since EEG signals exhibit large inter-subject variability and non-stationarity, EEG-based emotion recognition undergoes performance degradation across subjects and time~\cite{zheng2016personalizing}. Domain adaptation and generalization methods~\cite{siddhad2024enhancing}\cite{zhao2025federated} aim to align source and target distributions or extract invariant features, but their effectiveness is limited by substantial cross-subject differences. Recently, contrastive learning has shown promise by enhancing robustness through similarity learning. For example, SimCLR-based methods~\cite{chen2020simple} and the CLISA framework~\cite{shen2023contrastive} improved inter-subject alignment by enforcing consistent representations across subjects exposed to the same stimuli. However, it is challenging to construct effective positive/negative pairs in contrastive learning methods.

To overcome these limitations, we propose EmotionCLIP, which reformulates EEG-based emotion recognition as an EEG-text matching task. EmotionCLIP integrates a pre-trained CLIP text encoder~\cite{radford2021learning} with a new EEG backbone, SST-LegoViT, that sequentially captures spatial, spectral, and temporal features via multi-scale convolution, Legoformer modules, and Transformer encoders. By aligning EEG features with semantic text embeddings, the model improves cross-subject and cross-time generalization. Experiments on SEED~\cite{seed} and SEED-IV~\cite{seediv} datasets show that EmotionCLIP achieves accuracies of 88.69\% and 73.50\% in cross-subject recognition, and 88.46\% and 77.54\% in cross-time recognition, outperforming state-of-the-art baselines such as MSFR-GCN~\cite{msfrgcn}, MADA~\cite{MADA}, and CLISA~\cite{shen2023contrastive}.

The main contributions of this work are as follows:

 $\bullet$ We introduce the CLIP framework into EEG emotion recognition, enabling cross-domain generalization through EEG-text matching.

 $\bullet$ We design the SST-LegoViT, a lightweight backbone that learns EEG’s spatio-temporal-frequency features.

$\bullet$ Extensive experiments demonstrate that our method achieves superior performance on benchmark datasets, establishing a new state of the art for cross-domain recognition.

\begin{figure}[h]
    \centering
    \includegraphics[width=1\linewidth]{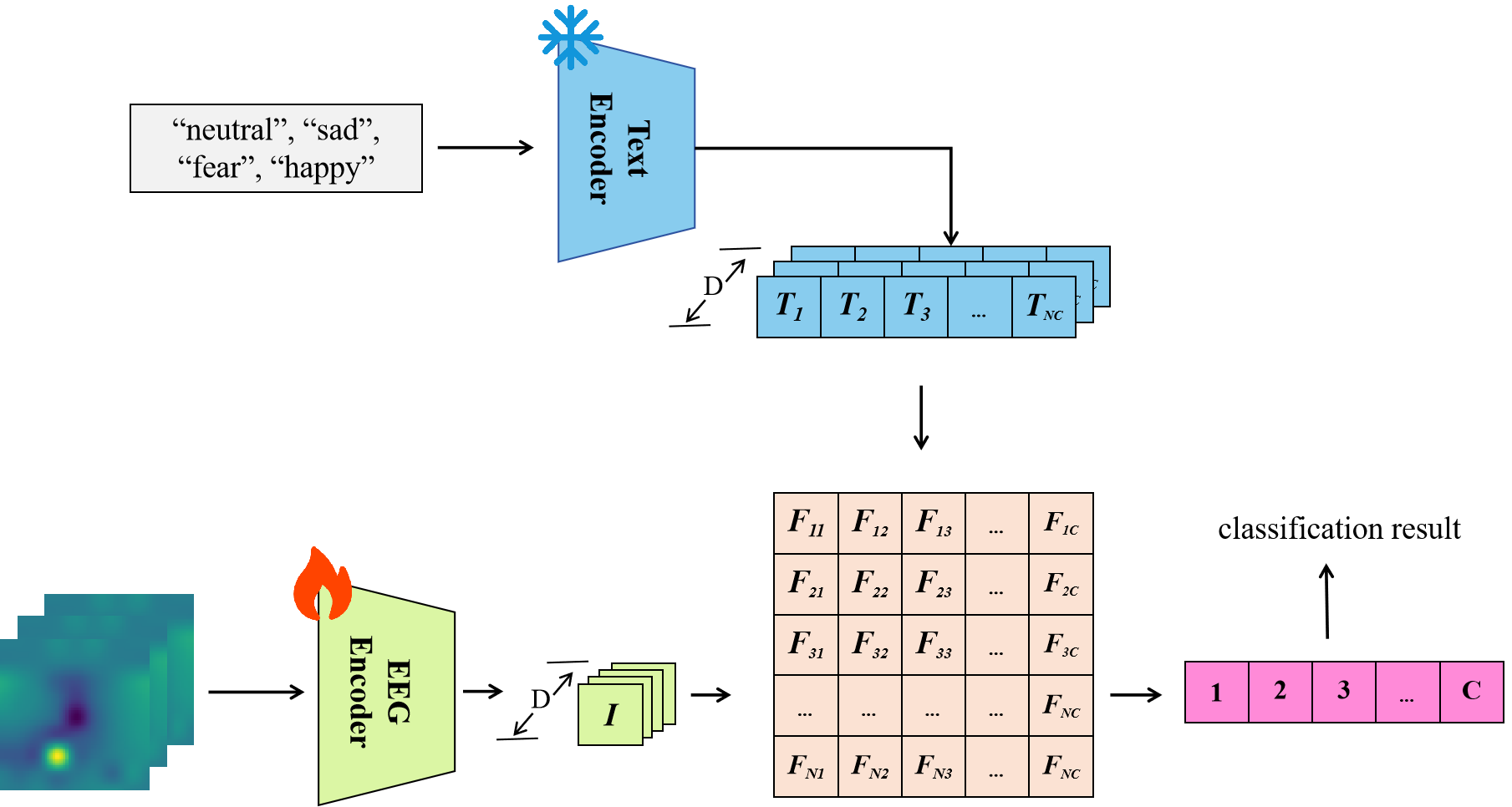}
    \caption{EmotionCLIP transforms the cross-domain emotion recognition task from a classification problem to an EEG-Text matching problem.}
    \label{fig:overall-architecher}
\end{figure}

\section{Methods}\label{sec:methods}
This section details the proposed EmotionCLIP model for cross-domain EEG-based emotion recognition. We first present the core research idea, followed by the overall architecture of the model, and then elaborate on the specifics of the Text Encoder and the novel EEG Encoder.

\subsection{EmotionCLIP}\label{architecture}

Traditional cross-domain methods, such as domain adaptation~\cite{siddhad2024enhancing}, aim to extract shared and invariant features directly from multi-source EEG data. However, this is challenging due to significant inter-subject variability and the non-stationary nature of EEG signals. Inspired by the success of the CLIP in multi-modal learning, we propose EmotionCLIP to reformulate the cross-domain emotion recognition task from a classification problem to an EEG-Text matching problem.

As shown in Figure~\ref{fig:overall-architecher}, EmotionCLIP introduces text as a stable intermediate modality, as textual descriptions of emotions are consistent across individuals and time.
The EmotionCLIP model employs a dual-encoder architecture, comprising a Text Encoder and an EEG Encoder. The model learns to align EEG signal features with the corresponding text embeddings of emotion labels. By maximizing the similarity between a given EEG sample and its correct emotional description.



\begin{figure}[h]
    \centering
    \includegraphics[width=1\linewidth]{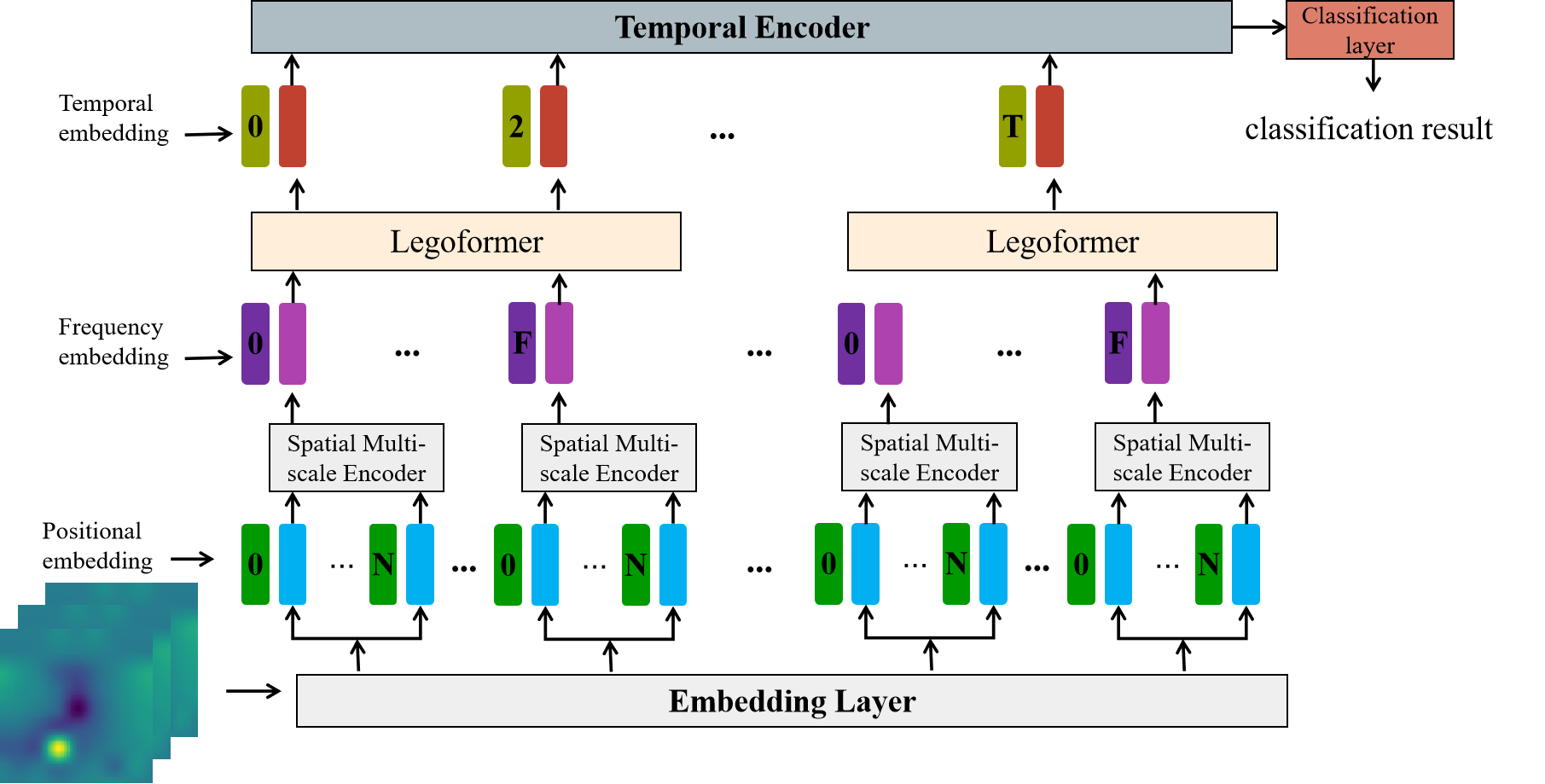}
    \caption{EEG Encoder sequentially processes spatial, frequency-band, and temporal information from a 4D EEG representation.}
    \label{fig:EEG-encoder}
\end{figure}

\textbf{Text Feature Generation:} Emotion labels (e.g., ``happy", ``sad") are inserted into predefined text prompt templates (e.g., ``The human feels {} now") to form descriptive sentences. These sentences are fed into the Text Encoder to generate a set of text feature embeddings.

\textbf{EEG Feature Generation:} Preprocessed 4D EEG data (Time × Frequency × Height × Width) is fed into our proposed SST-LegoViT model, which serves as the EEG Encoder, to produce a comprehensive EEG feature embedding that captures spatio-temporal-frequency information.

\textbf{EEG-Text Matching:} Both EEG and text features are projected into a shared embedding space. The model calculates the cosine similarity between the EEG feature and all text features. The emotion corresponding to the text feature with the highest similarity is selected as the prediction. The model is trained using a contrastive loss to maximize the similarity between correctly paired EEG-text features

\subsection{Text Encoder}\label{textencoder}

The Text Encoder is responsible for transforming emotion labels into rich semantic embeddings. We utilize the official pre-trained text encoder from the CLIP model. To provide richer context and mitigate ambiguity from single-word labels, we employ 16 text prompt templates. These templates are designed as declarative or interrogative sentences where the emotion label can be inserted, such as ``A video of {label} emotion" or ``The video makes the human feel $\{$label$\}$".

During training and inference, the parameters of the text encoder are kept frozen. This is a crucial step because the EEG datasets are relatively small, and fine-tuning the Text Encoder could disrupt the powerful, generalized semantic representations it learned from large-scale pre-training, thereby degrading performance.

\subsection{EEG Encoder}\label{eegencoder}

To effectively extract features from the complex, multi-dimensional EEG data, we designed a novel single-stream network named SST-LegoViT (Spatial-Spectral-Temporal Lego Vision Transformer).As shown in Figure~\ref{fig:EEG-encoder}, this model sequentially processes spatial, frequency-band, and temporal information from a 4D EEG representation.

\subsubsection{4D EEG Representation}
To leverage the complementary information in the spatio-temporal-frequency domains, we preprocess the raw EEG signals into a 4D tensor $X \in R^{T \times F \times H \times W}$.

First, the signal is filtered into multiple frequency bands (
$\delta \, [1{-}4 \, \text{Hz}]$,
$\theta \, [4{-}8 \, \text{Hz}]$,
$\alpha \, [8{-}14 \, \text{Hz}]$,
$\beta \, [14{-}31  \, \text{Hz}]$,
$\gamma_1 \, [31{-} \\ 51 \,  \text{Hz}]$ and
$\gamma_2 \, [51{-}75 \, \text{Hz}]$). For each band, we extract Differential Entropy (DE)~\cite{de} and Power Spectral Density (PSD)~\cite{psd} features.

Next, to capture spatial information, the features from the 62 channels are mapped onto a 2D grid based on their electrode positions, creating a sparse image-like representation. This grid is then up-sampled via interpolation to a higher resolution (e.g., 64×64) to enhance spatial details.

Finally, the signal is segmented in time to create multiple frames, forming the temporal dimension.

\subsubsection{SST-LegoViT Architecture}
The SST-LegoViT model consists of four main modules: an Embedding Layer, a Spatial Multi-scale Encoder, a Frequency-band Encoder (Legoformer), and a Temporal Encoder.

\textbf{Embedding Layer:} Instead of the standard ViT's single large-kernel convolution for patching, we use a stack of four smaller 2D convolutional layers. This approach creates more stable patch embeddings from the 2D spatial maps of the EEG data for each time step and frequency band.

\begin{figure}[t]
    \centering
    \includegraphics[width=1\linewidth]{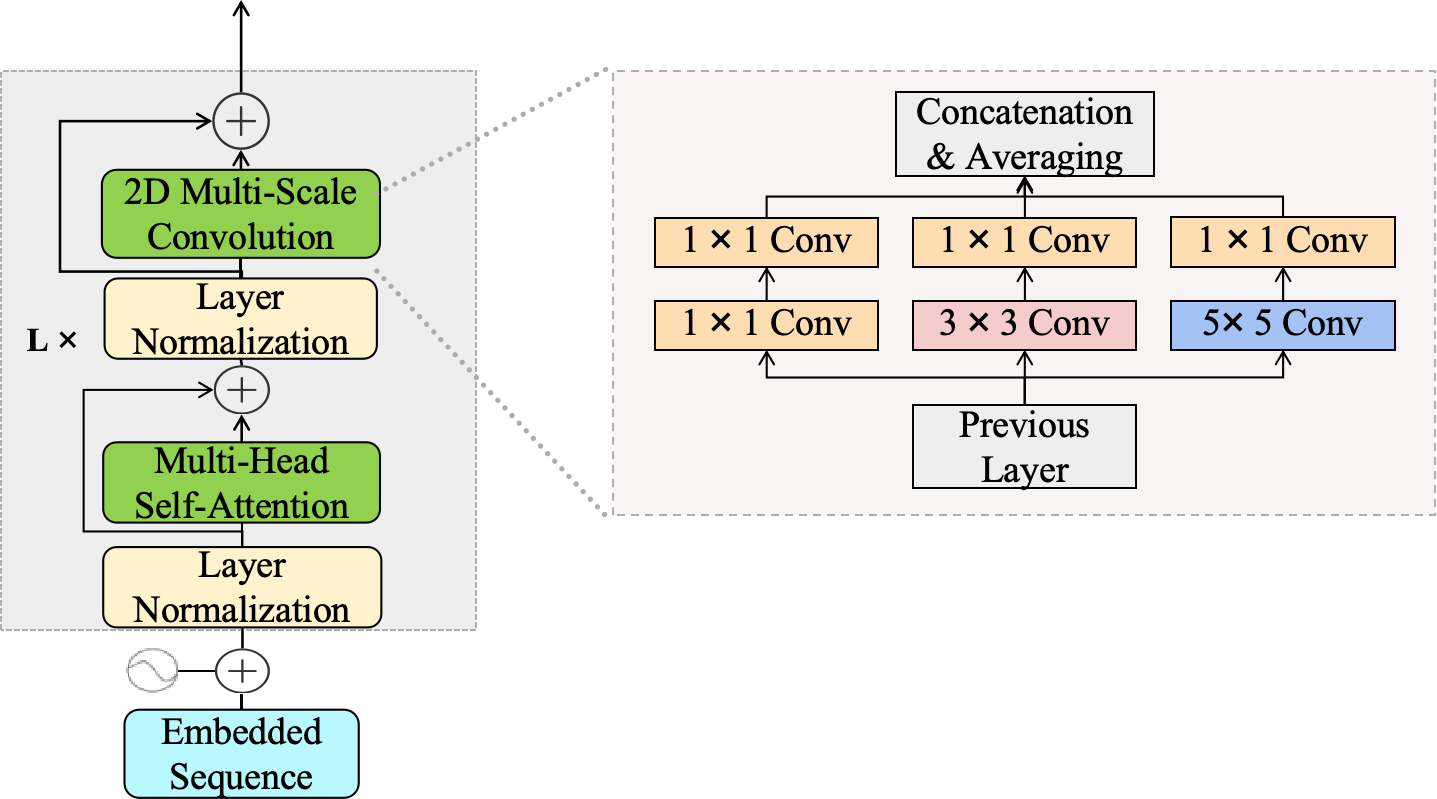}
    \caption{Spatial Multi-scale Encoder aggregates multi-resolution features to enhance spatial representation across varying scales.}
    \label{fig:spatialEncoder}
\end{figure}

\textbf{Spatial Multi-scale Encoder:} To capture spatial features at various scales, we modify the standard Transformer block. As shown in Figure~\ref{fig:spatialEncoder}, the feed-forward network (FFN) is replaced with a 2D multi-scale convolutional module, which contains parallel branches with 1$\times$1, 3$\times$3, and 5$\times$5 kernels. The outputs of these branches are aggregated to produce a multi-scale spatial representation.

\begin{figure}[t]
    \centering
    \includegraphics[width=1\linewidth]{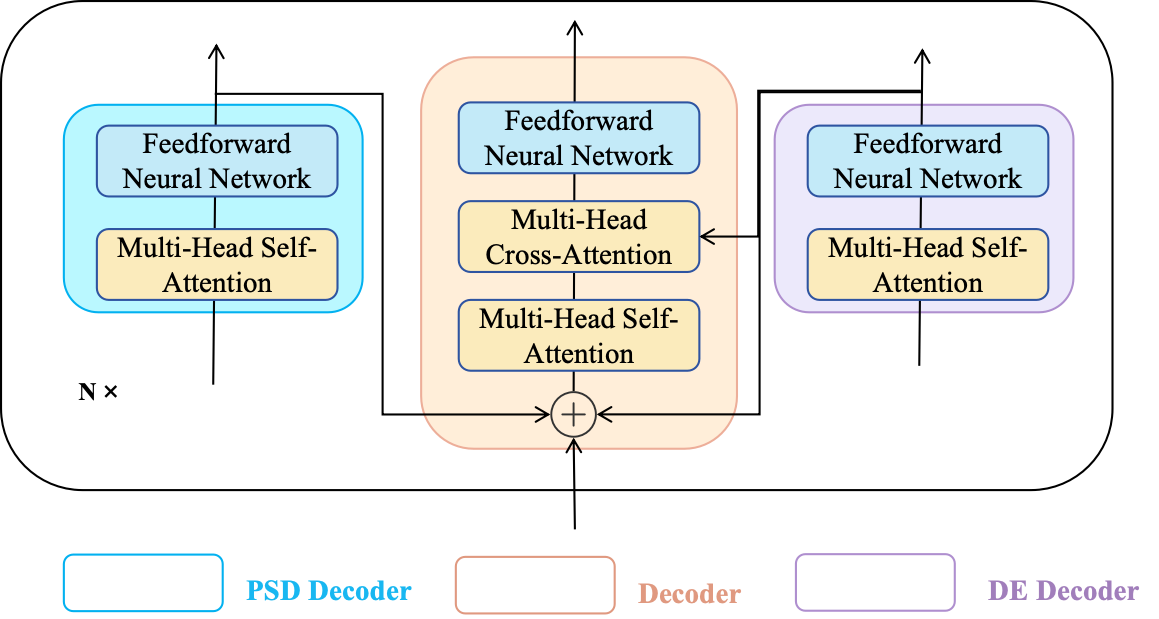}
    \caption{LegoFormer processes DE and PSD features through parallel encoders. Fuses them via cross-attention, using DE as the primary context to guide auxiliary PSD information.}
    \label{fig:legoformer}
\end{figure}

\textbf{Frequency-band Encoder (Legoformer):} Different frequency-domain features (DE~\cite{de} and PSD~\cite{psd}) contribute differently to emotion recognition. To effectively and adaptively fuse them, we introduce the Legoformer module. As shown in Figure~\ref{fig:legoformer}, This module consists of separate, parallel Transformer encoders for each feature type (one for DE, one for PSD). The outputs are then fused and fed into a decoder module, which uses cross-attention with the primary feature's (DE) output as the key and value. This allows the model to leverage auxiliary features (PSD) while retaining focus on the most informative primary features (DE).

\textbf{Temporal Encoder:} After feature fusion across the frequency-band dimension, the resulting sequence of features across time frames is processed by a standard Transformer encoder. This final module captures the temporal dynamics and long-range dependencies within the EEG signal to produce the final EEG feature embedding.

\section{results}\label{sec:results}

\subsection{Datasets and Experimental Setup}\label{datasetAndSetup}
All experiments were conducted on two publicly available EEG emotion datasets, SEED~\cite{seed} and SEED-IV~\cite{seediv}. To evaluate the model’s cross-domain generalization ability, two rigorous evaluation strategies were adopted: cross-subject validation using the leave-one-subject-out (LOSO) protocol, and cross-time validation across different sessions of the same subject. The model was trained with the AdamW optimizer, a batch size of 64, an initial learning rate of 0.0001 with a cosine decay schedule, and a weight decay of 0.003. Cross-entropy loss was employed, and an early stopping strategy was applied to terminate training if the validation accuracy did not improve for 50 consecutive epochs.




\subsection{Experimental Results}\label{RxperimentResults}
\subsubsection{Zero-shot and Few-shot Results}
\begin{figure}[t]
    \centering
    \captionsetup[subfigure]{font=tiny} 
    \begin{subfigure}[b]{0.235\textwidth}
        \includegraphics[width=\textwidth]{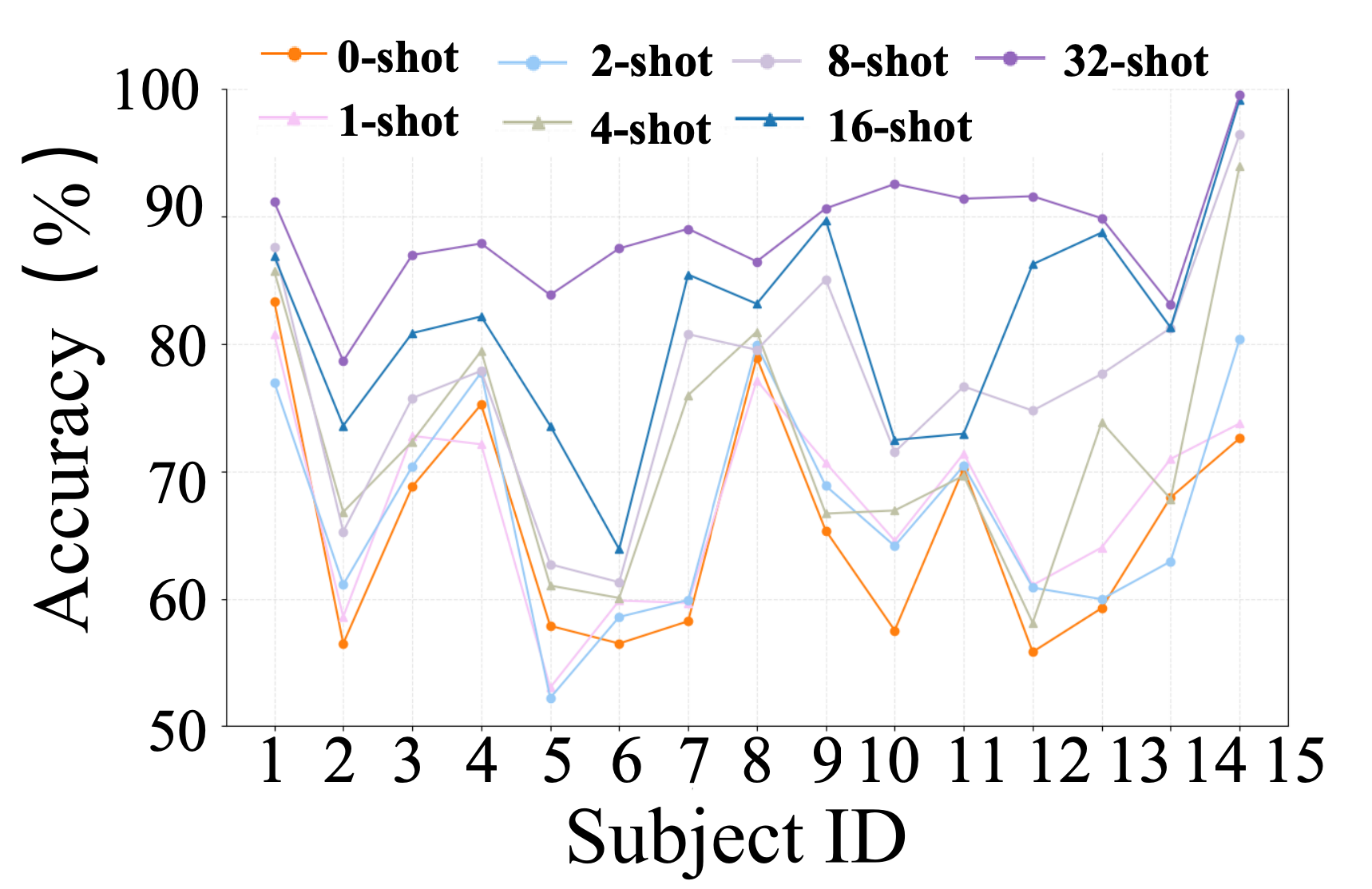}
        \caption{Few-Shot Results on SEED Dataset}
        \label{fig:sub1}
    \end{subfigure}
    \begin{subfigure}[b]{0.235\textwidth}
        \includegraphics[width=\textwidth]{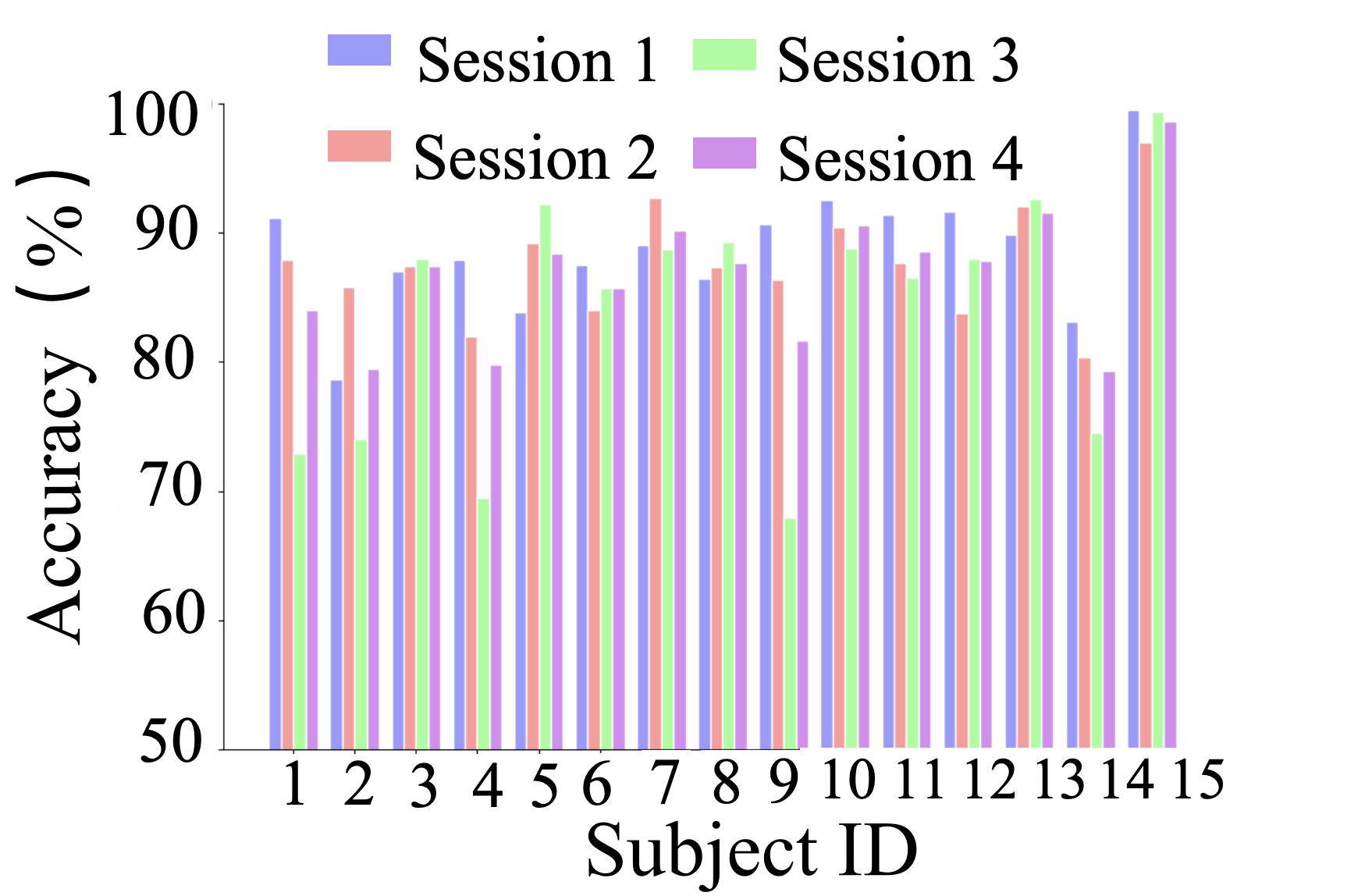}
        \caption{Cross-Subject Strategy Results on SEED Dataset}
        \label{fig:sub2}
    \end{subfigure}
    \begin{subfigure}[b]{0.235\textwidth}
        \includegraphics[width=\textwidth]{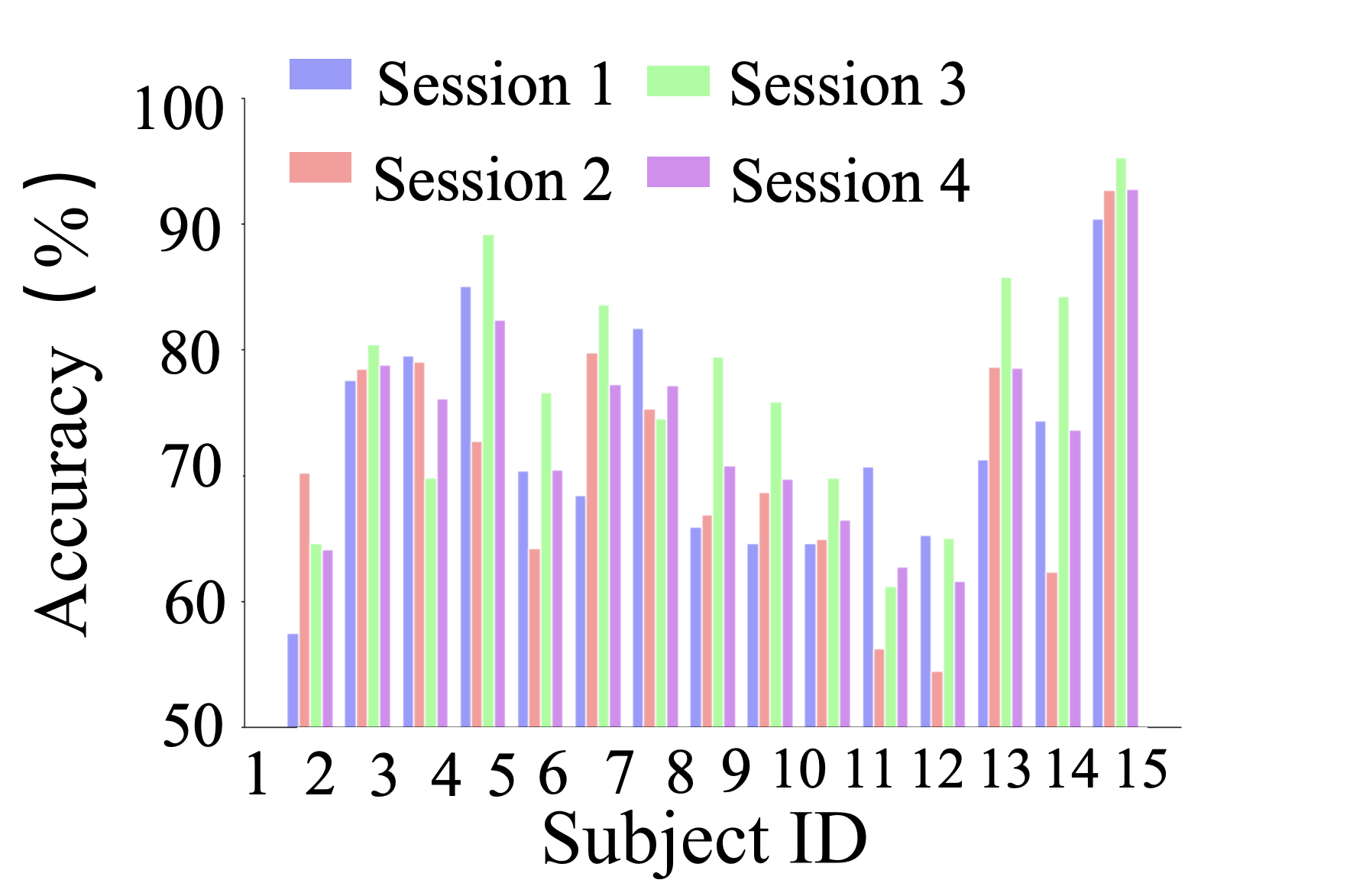}
        \caption{Cross-Subject Results on SEED-IV Dataset}
        \label{fig:sub3}
    \end{subfigure}
    \begin{subfigure}[b]{0.235\textwidth}
        \includegraphics[width=\textwidth]{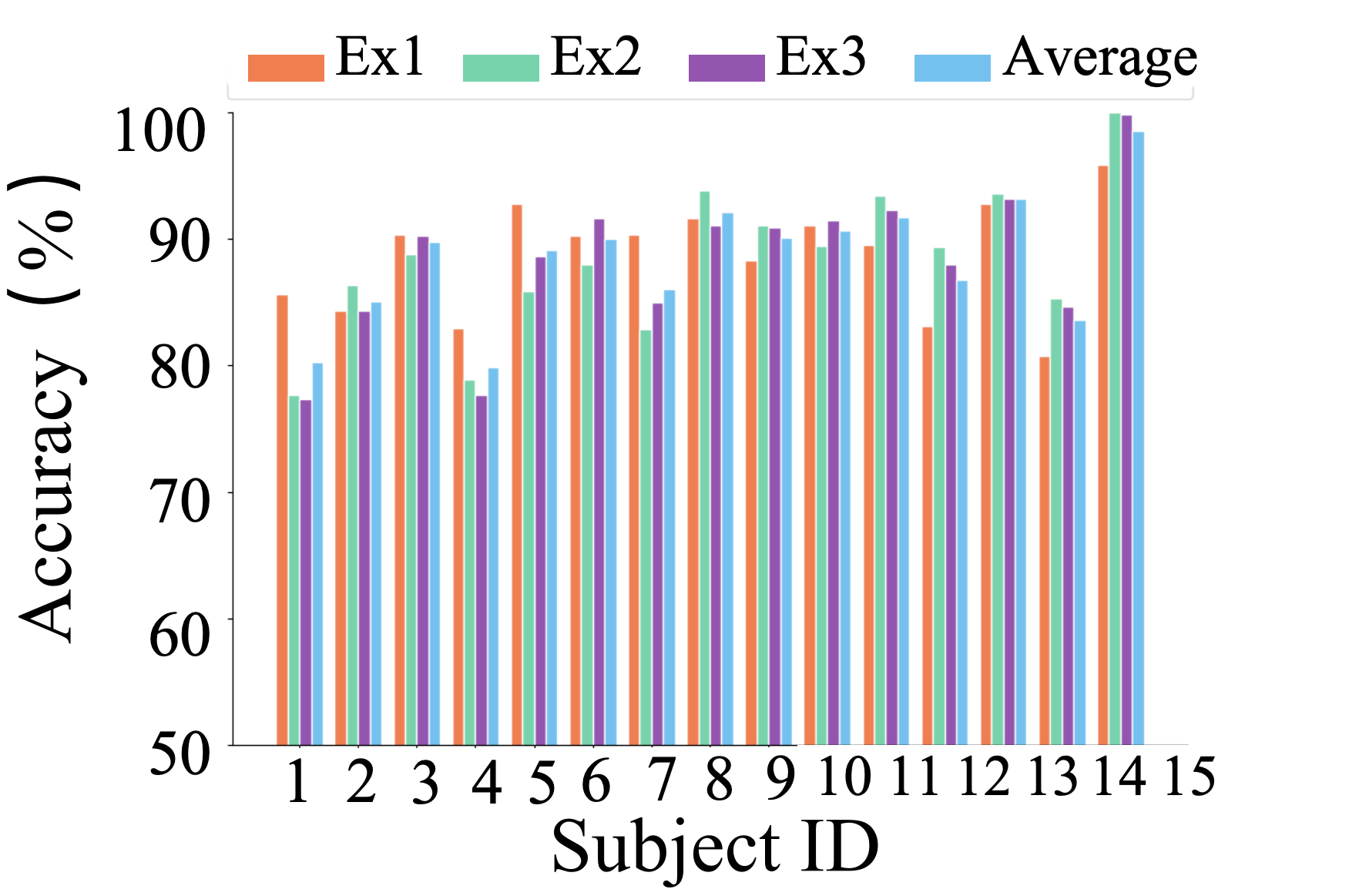}
        \caption{Cross-Time Results on SEED Dataset}
        \label{fig:sub4}
    \end{subfigure}

    \begin{subfigure}[b]{0.235\textwidth}
        \includegraphics[width=\textwidth]{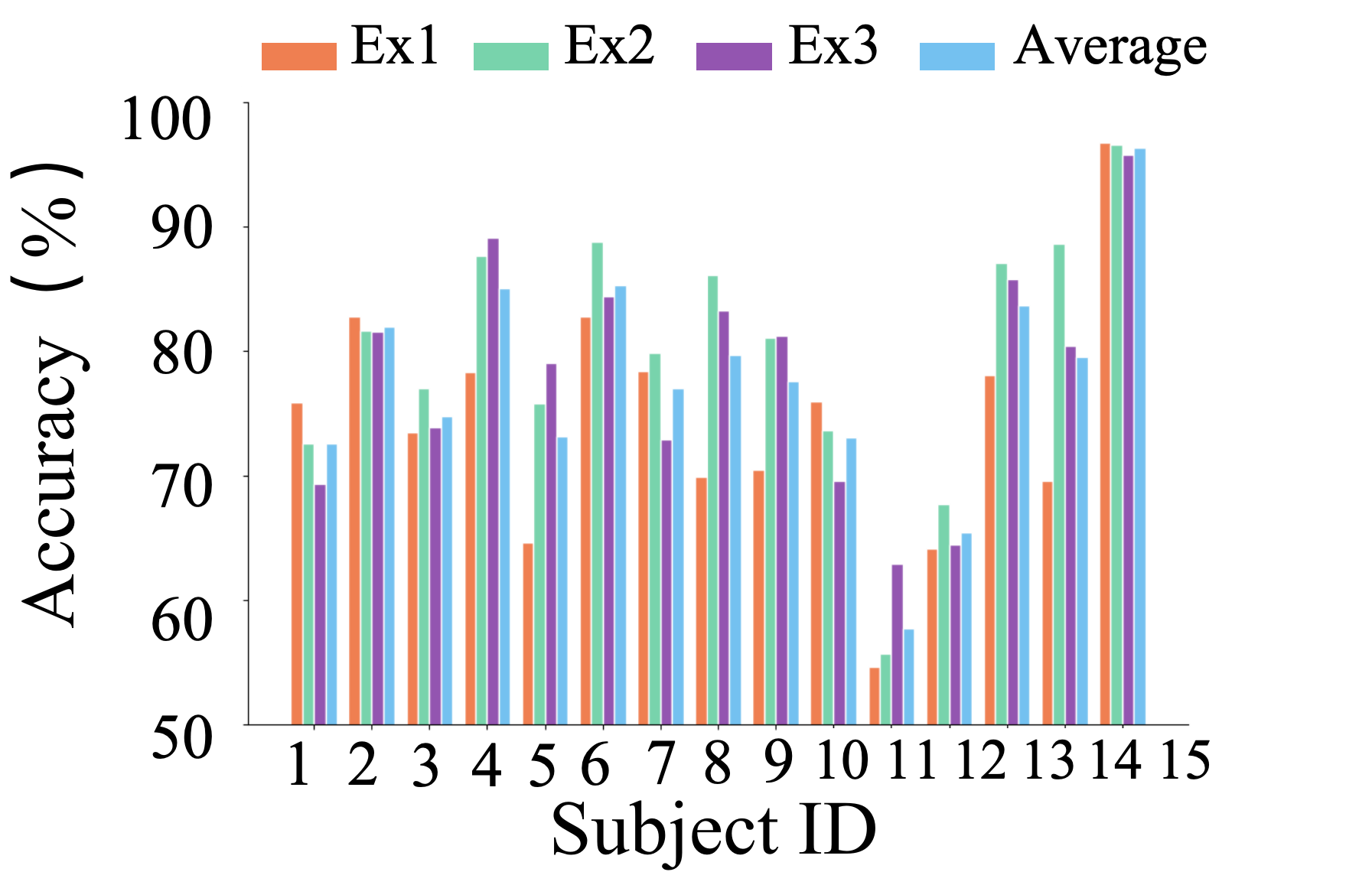}
        \caption{Cross-Time Results on SEED-IV Dataset}
        \label{fig:sub5}
    \end{subfigure}
    \begin{subfigure}[b]{0.235\textwidth}
        \includegraphics[width=\textwidth]{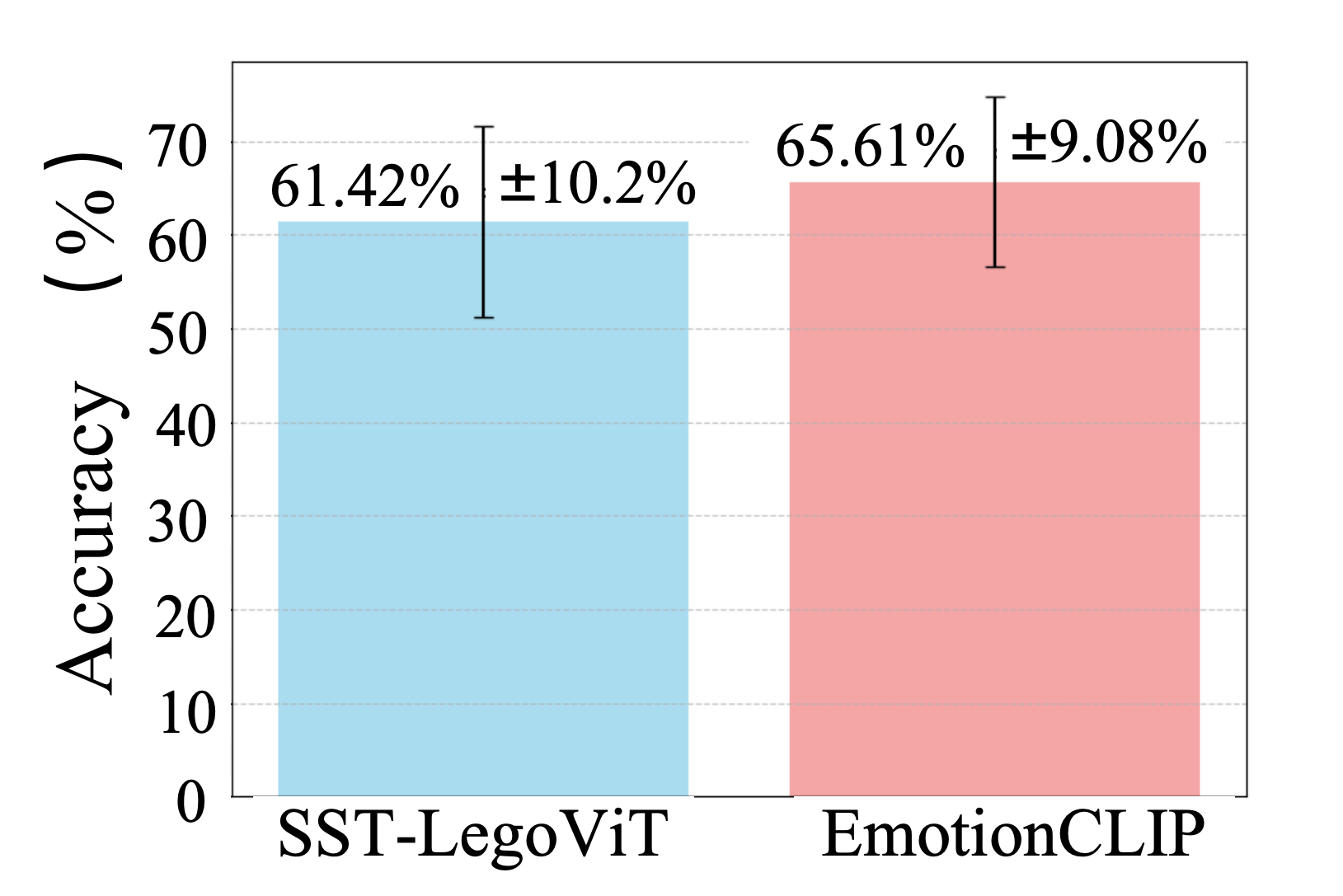}
        \caption{Ablation Results under Cross-Subject Strategy}
        \label{fig:sub6}
    \end{subfigure}
    
    \caption{Experimental Results.}
    \label{fig:total}
    \vspace{-16pt}
\end{figure}

While CLIP offers zero-shot capability, its performance declines on abstract tasks with distributions differing from pre-training. We evaluated EmotionCLIP on SEED under zero-shot and N-shot settings. As shown in Fig. ~\ref{fig:sub1}, accuracy improves consistently as N increases from 0 to 32. In the zero-shot case, performance varies widely across subjects, with some below 60\%, whereas few-shot training significantly boosts accuracy. This suggests that limited domain-specific fine-tuning is crucial for EEG emotion recognition. All subsequent experiments adopt the 32-shot setting, denoted as EmotionCLIP-32.

\begin{table}[t]
\centering
\scriptsize
\caption{Comparison with state-of-the-art models on cross-subject emotion recognition  }
\label{tab:compare}
\adjustbox{max width=\textwidth}{
\begin{tabular}{l c c c c}
\toprule
\multirow{2}{*}{Model} & \multicolumn{2}{c}{SEED} & \multicolumn{2}{c}{SEED-IV} \\
\cmidrule(lr){2-3} \cmidrule(lr){4-5}
 & ACC(\%) & STD(\%) & ACC(\%) & STD(\%) \\
\midrule
TPT~\cite{zheng2016personalizing}       & 76.31  & 15.89 & $\backslash$ & $\backslash$ \\
DGCNN~\cite{song2018dgcnn}      & 79.95  & 9.02  & 52.82      & 9.23      \\
BiHDM~\cite{li2020bi}      & 85.40  & 7.53  & 69.03      & 8.66      \\
RGNN~\cite{zhong2020rgnn}     & 85.3   & 6.72  & 73.84      & 8.02      \\
CLISA~\cite{shen2023contrastive}      & 86.4   & 6.4   & $\backslash$ & $\backslash$ \\
MADA~\cite{MADA}      & 86.16  & 7.87  & 59.29      & 13.65     \\
MSFR-GCN~\cite{msfrgcn}    & 86.78  & 5.40  & 73.43      & 7.32      \\
\textbf{EmotionCLIP-32}        & 88.69  & 4.82  & 73.50      & 9.73      \\
\bottomrule
\end{tabular}
}
\vspace{-10pt}
\end{table}

 \vspace{-6pt}
\subsubsection{Comparison with Baseline Models}
We compared EmotionCLIP-32 with state-of-the-art methods for cross-subject EEG emotion recognition (Table ~\ref{tab:compare}). On SEED, it achieves 88.69\% accuracy, surpassing MSFR-GCN by 1.91\% and MADA by over 2\%. On SEED-IV, it attains 73.50\%, slightly below RGNN but higher than MSFR-GCN. These results highlight the effectiveness of our EEG-text matching framework.

\vspace{-6pt}
\subsubsection{Cross-Subject Results}



We assessed cross-subject performance on both SEED and SEED-IV datasets, with 15 subjects completing three sessions each. Leave-one-subject-out cross-validation was performed per session, and final accuracy was averaged over all subjects and sessions.

On SEED, average accuracies were 88.69\%, 87.59\%, and 83.87\%; on SEED-IV, they were 72.27\%, 70.98\%, and 77.04\%. As shown in Fig. ~\ref{fig:sub2}~\ref{fig:sub3}, accuracy varied considerably across subjects (e.g., Subject 15 \textgreater{98\%}, Subjects 4 and 9 lower), highlighting individual EEG differences. Nevertheless, our model achieved strong average performance, demonstrating robustness.

\subsubsection{ Cross-Time Results}

We evaluated temporal stability through three cross-session experiments (Table ~\ref{tab:cross-time-results}). On SEED, EmotionCLIP-32 maintains an average accuracy of 88.46\%, and on SEED-IV, 77.54\%. As shown in Fig. ~\ref{fig:sub4}, \ref{fig:sub5}, most subjects perform consistently well (e.g., Subject 15), while a few (e.g., Subject 11) show lower stability, suggesting greater variability in their EEG patterns.

\begin{table}[t]
\centering
\scriptsize
\caption{Table 2 Dataset division methods and experimental results under cross-time strategy}
\label{tab:cross-time-results}
\adjustbox{max width=\textwidth}{
\begin{tabular}{c c c c c c c}
\toprule
\multirow{2}{*}{No.} & \multirow{2}{*}{Training Set} & \multirow{2}{*}{Test Set} & \multicolumn{2}{c}{SEED} & \multicolumn{2}{c}{SEED-IV} \\
\cmidrule(lr){4-5} \cmidrule(lr){6-7}
 & & & ACC(\%) & STD(\%) & ACC(\%) & STD(\%) \\
\midrule
Ex 1 & Session 1 & Session 2 & 88.65 & 4.19 & 74.38 & 9.46 \\
Ex 2 & Session 1 & Session 3 & 88.31 & 5.66 & 79.99 & 9.81 \\
Ex 3 & Session 2 & Session 3 & 88.42 & 5.68 & 78.24 & 8.94 \\
\bottomrule
\end{tabular}
}
\end{table}

\subsubsection{Ablation Study Results}

We performed an ablation study on SEED under the cross-subject protocol. As shown in Figure~\ref{fig:sub6}, the SST-LegoViT encoder alone achieved 61.42\% accuracy, while adding the text encoder and reframing the task as EEG-text matching raised zero-shot EmotionCLIP to 65.61\% (+4.19\%). This highlights the effectiveness of aligning EEG with semantic text space for cross-domain generalization.


\section{Conclusion}\label{sec:conclusion}

We introduced EmotionCLIP, a novel framework that significantly enhances cross-domain EEG emotion recognition by reformulating the task as an EEG-text matching problem. Our approach utilizes the proposed SST-LegoViT architecture to extract comprehensive spatio-temporal-frequency features from EEG signals effectively. Extensive experiments on the SEED and SEED-IV datasets show that EmotionCLIP achieves state-of-the-art performance in challenging cross-subject and cross-time evaluations. These results confirm that aligning EEG features with a stable, multi-modal semantic space is a powerful and effective paradigm for building robust and generalizable affective computing systems.

\clearpage

{
\bibliographystyle{IEEEbib}
\bibliography{strings,refs}
}
\end{document}